\documentclass{bmvc2k}

\usepackage{amsfonts}
\usepackage{booktabs}
\usepackage{color}
\usepackage{epstopdf}
\newcommand{\jfh}[1]{{\color{black}{#1}}}

\newcommand{\bing}[1]{{\color{black}{#1}}}

\title{Improving Fast Segmentation With Teacher-student Learning}

\addauthor{Jiafeng Xie}{xiejf6@mail2.sysu.edu.cn}{1}
\addauthor{Bing Shuai}{bshuai001@ntu.edu.sg}{2}
\addauthor{Jian-Fang Hu$^{\text {1,3,}}$}{hujf5@mail.sysu.edu.cn}{4}
\addauthor{Jingyang Lin}{linjy47@mail2.sysu.edu.cn}{1}
\addauthor{Wei-Shi Zheng$^{\text {1,}}$}{wszheng@ieee.org}{4}

\addinstitution{
School of Data and Computer Science, Sun Yat-sen University, China
}
\addinstitution{
 Nanyang Technological University, Singapore
}
\addinstitution{Guangdong Key Laboratory of Information Security Technology}
\addinstitution{Key Laboratory of Machine Intelligence and Advanced Computing, MOE}

\runninghead{XIE et. al}{IMPROVING FAST SEGMENTATION WITH TEACHER-STUDENT LEARNING}

\begin{document}

\maketitle

\begin{abstract}
Recently, segmentation neural networks have been significantly improved by demonstrating very promising accuracies on public benchmarks. However, these models are very heavy and generally suffer from low inference speed, which limits their application scenarios in practice. Meanwhile, existing fast segmentation models usually fail to obtain satisfactory segmentation accuracies on public benchmarks. In this paper, we propose a teacher-student learning framework that transfers the knowledge gained by a heavy and better performed segmentation network (i.e. teacher)  to guide the learning of fast segmentation networks (i.e. student). Specifically, both zero-order and first-order knowledge depicted in the fine annotated images and unlabeled auxiliary data are transferred to regularize our student learning. The proposed method can improve existing fast segmentation models without incurring extra computational overhead, so it can still process images with the same fast speed. Extensive experiments on the Pascal Context, Cityscape and VOC 2012 datasets demonstrate that the proposed teacher-student learning framework is able to significantly boost the performance of student network.

\end{abstract}

%-------------------------------------------------------------------------
\section{Introduction}
\label{sec:intro}
%Semantic segmentation has drawn more and more attention in the past few decades ~\cite{long2015fully,chen2018deeplab,zhao2017p                                                         yramid};automation robotics ~\cite{nycz2018additive} and augmented reality ~\cite{santos2016augmented} etc. Segmentation accuracy as well as speed are two principle protocols used for evaluating segmentation models and the existing methods fall into two main groups: high-fidelity segmentation maps ~\cite{long2015fully,chen2018deeplab,zhao2017pyramid,ronneberger2015u} as well as efficient inference ~\cite{zhao2017icnet,paszke2016enet,badrinarayanan2017segnet}.

\jfh{Recently,} segmentation performance has been lamdrastically improved in deep learning era, where end-to-end segmentation networks~\cite{long2015fully,chen2018deeplab,zhao2017pyramid} are developed to generate high-fidelity segmentation maps on challenging real-world images ~\cite{everingham2012pascal,cordts2016cityscapes,mottaghi2014role}. For example, deeper and higher-capacity ConvNets ~\cite{long2015fully,zheng2015conditional,he2016deep} are adopted in Fully Convolutional Network (FCN) to enhance its segmentation accuracies.   Some researchers are dedicated to aggregating informative contexts for local feature representation, which leads to advancement of segmentation network architectures for boosting the segmentation accuracies. Meanwhile, a large body of research works focus on refining local segmentation results to obtain non-trivial performance enhancement of segmentation networks. Some representative works include DeepLab-v2~\cite{chen2018deeplab}, DilatedNet~\cite{yu2015multi}, CRF-CNN~\cite{lin2016efficient}, DAG-RNN~\cite{shuai2016dag}, PSPNet~\cite{zhao2017pyramid}, etc. In addition,  some researchers also explore refining the low-level segmentation details (e.g. sharper object boundaries and better segmentation accuracies for small-size objects) by either adopting  fully connected CRF as a post-processing module ~\cite{chen2018deeplab,zheng2015conditional} or by learning a deconvolutional network on top of coarse prediction maps ~\cite{noh2015learning}. It's interesting to see that segmentation accuracies are progressively improved on public benchmarks, however, these models become less likely to be applicable in real-world scenarios where short latency is essential and computational resources are limited.

With the above concerns in mind, recent researchers start to ingest the recent advancement of network architectures, and integrate them to develop faster segmentation networks. Representative examples are ENet ~\cite{paszke2016enet}
and SegNet~\cite{badrinarayanan2017segnet}. Although promising, their segmentation accuracies are still bounded by the model capacity. In general, they fail to obtain comparable segmentation accuracies compared to heavy and deep segmentation networks. In this paper, we propose a novel framework to improve the performance of fast segmentation network without incorporating extra model parameters or incurring extra computational overhead, and thus can keep the inference speed of the fast segmentation network to be unchanged. To this end, we propose a novel teacher-student learning framework to make use of the knowledge gained in a teacher network. Specifically, our framework intend to regularize the student learning by the zero-order and first-order knowledge obtained from teacher network on fine annotated training data. To distill more knowledge from teacher, we further extend our framework by integrating the teacher-student learning on fine annotated training data and unlabeled auxiliary data. Our experiments show that the proposed teacher-student learning framework can boost the performance of student network by a large margin.

%With the recent progress of deep learning techniques \cite{Simonyan2014Very} \cite{Krizhevsky2012ImageNet}, high quality segmentation approaches intend to segment images with high accuracy by employing very deep and complex networks \cite{Long2014Fully,Chen2014Semantic,Zhao2017Pyramid}. However, these models suffer from a low segmentation speed (e.g., 5 FPS for Resnet-DeeplabV2), which makes them less applicable in some real-world applications. In contrast, the methods for fast segmentation can segment images with a fast speed by using a light weight network \cite{Badrinarayanan2017SegNet,Paszke2016ENet} or simplified high quality segmentation model \cite{Zhao2017ICNet} (Double checked  ICNet???). However, compared with the high quality segmentation models, these fast segmentation models are more easily to fall into local optimal solutions and thus can not learn enough knowledge to achieve good segmentation results. Recent researches

Our contribution is three-fold: (1) a novel teacher-student learning framework for improving fast segmentation models, with the help of an auxiliary teacher network; (2) a joint learning framework for distilling knowledge from teacher network based on both fine annotated data and unlabeled data; and (3) extensive experiments on three segmentation datasets demonstrating the effectiveness of the proposed method.

\section{Related work}

In the following, we review the literatures that are most relevant to our work, including researches of architecture evolvement for semantic segmentation and knowledge distillation.

%In the following, we briefly review the related work of our proposed methods. We first introduce the methods that focus on improving segmentation accuracy,which are the teacher network in our methods, followed by the methods that focus on improving segmentation speed, which are the student network in our methods. Then we outline the knowledge distillation technique which transfers knowledge between networks, and the MobileNet\cite{ref7} which are used as a basic feature extraction module in our MobileNet-deeplabV2\cite{ref2}\cite{ref7} network.
\noindent \textbf{Accuracy Oriented Semantic Segmentation.} This line of research covers most of published literatures in semantic segmentation. In a nutshell, the goal is to significantly improve the segmentation accuracy on public segmentation benchmarks. Following the definition of a general segmentation network architecture in Shuai et al. \cite{shuai2017scene}, we categorize the literatures to three aspects that improve the segmentation performance.
In one aspect, performance enhancement is largely attributed to the magnificent progress of pre-trained ConvNet \cite{krizhevsky2012imagenet}  \cite{szegedy2015going} \cite{simonyan2014very}\cite{he2016deep}  \cite{huang2017densely}, which is simply adapted to be the local feature extractor in segmentation networks. The core of this progress is to obtain better ConvNet model on large-scale image datasets (e.g. ImageNet \cite{russakovsky2015imagenet} ) by training deeper or more complicated networks. Meanwhile, many researchers are dedicated to developing novel computational layers that are able to effectively encode informative context into local feature maps. This research direction plays a significant role to enhance the visual quality of prediction label maps as well as to boost the segmentation accuracy. Representative works, such as DPN \cite{liu2015semantic}, CRF-CNN  \cite{lin2016efficient} , DAG-RNN  \cite{shuai2017scene}, DeepLab-v2 \cite{chen2018deeplab} , RecursiveNet \cite{sharma2014recursive}, ParseNet \cite{liu2015parsenet}, DilatedNet \cite{yu2015multi} , RefineNet \cite{lin2017refinenet}, PSPNet \cite{zhao2017pyramid}  formulated their computational layers to achieve effective context aggregation, and they can significantly improve the segmentation accuracy on Pascal VOC benchmarks. In addition, research endeavours have also been devoted to recovering the detailed spatial information by either learning a deep decoder network \cite{badrinarayanan2017segnet}\cite{noh2015learning} \cite{wang2017understanding}  or applying a disjoint post-processing module such as fully connected CRF \cite{krahenbuhl2011efficient} \cite{chen2018deeplab}\cite{zheng2015conditional}.
These techniques have collectively pushed the segmentation performance saturating on Pascal VOC benchmarks\footnote{ More than 85\% mean IOU has been achieved by state-of-the-art segmentation models.}. The steady progress also calls for the unveiling of new and more challenging benchmarks (e.g. Microsoft COCO \cite{lin2014microsoft} dataset). Although significant progress has been made regarding to the visual quality of segmentation predictions, these models are usually computationally intensive. Thus, they are problematic to be directly applied to resource constrained embedded devices and can not be used for real-time applications.

\noindent \textbf{Fast Semantic Segmentation.}  Recently, another line of research emerges as of state-of-the-art models achieve saturating segmentation accuracies on urban street images \cite{cordts2016cityscapes}. Its goal is to develop fast segmentation models that has the potential to be applied in real-world scenarios. For example,  Paszke et al.\cite{paszke2016enet} adopted  a light local feature extraction network  in their proposed ENet, which can be run in real-time for moderate sized images (e.g. 500 x 500). Zhao et al. \cite{zhao2017icnet} developed the ICNet that only fed the heavy model with downsampled input images, so the inference speed of ICNet remains competitively fast. One problem of these works are that the performance of these models are not satisfactory due to their lower capacity. In this paper, we propose to improve the performance of fast segmentation networks by regularizing their learning with the knowledge learned by a heavy and accurate teacher model. In this regard, this line of research is orthogonal and complementary to our teacher-student learning framework.

%One of the popular way to improve segmentation speed is to reduce the number of model parameters. This is achieved by discarding the deconvolution layers \cite{badrinarayanan2017segnet}, using light deep neural networks \cite{paszke2016enet}, reducing the input resolution of the deepest network \cite{zhao2017icnet}. Although these methods can largely increase computation efficiency and obtain fast segmentation, the segmentation performance suffers a lot.

\vspace{0.1cm}
\noindent \textbf{Knowledge Distillation via Teacher-Student Learning.} In image classification community, knowledge distillation \cite{gulccehre2016knowledge, hinton2015distilling, zagoruyko2016paying,huang2017like,yim2017gift}  has been widely adopted to improve the performance of fast and low-capacity neural networks.  Hinton et al.
\cite{hinton2015distilling} pioneered to propose transfer the ``dark knowledge'' from an ensemble of networks to a student network, which leads to a significant performance enhancement on ImageNet 1K classification task \cite{russakovsky2015imagenet}.  Romero et al. \cite{romero2014fitnets} further extended the knowledge transfer framework to allow it happens in intermediate feature maps. Their proposed  FitNet\cite{romero2014fitnets}  allowed the architecture of student network to go deeper and more importantly to achieve better performance.
Huang et al. \cite{huang2017like} proposed to regularize the student network learning by mimicking the distribution of activations of intermediate layers in a teacher network. In addition, knowledge distillation has also been sucessfully in pedestrian detection \cite{hu2017pushing} and face recognition \cite{tai2016face} as well.  Recently, Ros et al. \cite{ros2016training} explored and discussed different knowledge transfer framework based on the output probability of a teacher deconvolutional network, and they observed segmentation accuracy improvement of student networks. Our methods differ from it  in the following aspects: (1) both zero-order and first-order knowledge from teacher models are transferred to student; and (2) unlablled auxiliary data are used to encode the knowledge of teacher models, which is further transferred to the student models and improve their performance.

% \begin{figure*}[htb]
% \begin{center}
% %\fbox{\rule{0pt}{2in} \rule{.9\linewidth}{0pt}}
% \includegraphics[width=0.9\linewidth]{images/motivation2.png}
% \end{center}
%    \caption{The explanation for consistency map and the purpose of our proposed method.(a) Show the input image and the consistency map which mainly focus on the the correlation of the adjacent pixels.(b) Our proposed method aim to transfer the probability and consistency knowledge from the teacher network to the student network.}
% \label{fig:comb}
% \end{figure*}

\section{Approach} \label{Section:Approach}

\begin{figure*}[t]
\begin{center}
%\fbox{\rule{0pt}{2in} \rule{.9\linewidth}{0pt}}
\includegraphics[width=0.8\linewidth, height= 0.35\linewidth]{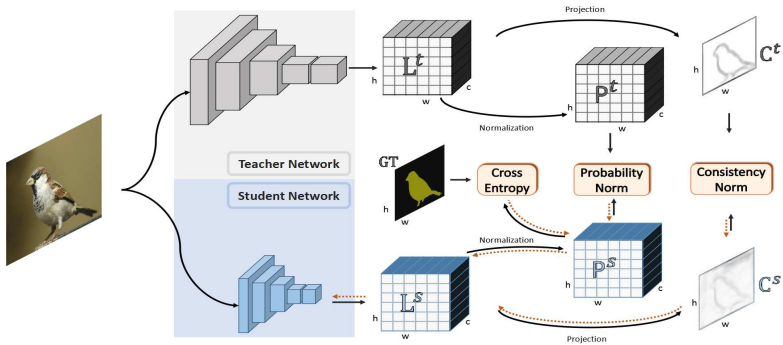}
\end{center}
\vspace{-0.4cm}
   \caption{The detailed architecture of our teacher-student learning framework. In the framework optimization, only the student networks are updated by back-propagation (indicated by red dashed lines) and the teacher network are fixed. For transferring zero-order and first-order knowledge, we generate the probability maps and consistency maps using the logits output of the teacher and student networks.}
\label{fig:Architecture}
\end{figure*}

Our approach involves two kind of deep networks: student network and teacher network. The student network is a deep network for segmentation with a shallower architecture. Thus it can segment images with a fast speed. In contrast, the teacher network is a deeper network with more complex architectures. Thus, it typically performs better than the student network in the term of segmentation accuracy, but has a slower segmentation speed. In this work, we propose a teacher-student learning framework to improve the student learning with the guidance of a teacher network. The proposed overall framework is summarized in Figure \ref{fig:Architecture}. In the following, we discuss it in detail.

%\subsection{Revisit to the  teacher-student learning with Logit Regression}

\subsection{Teacher-student learning for fine annotated data}

Here, we describe how to facilitate the learning of student network with the help of a teacher network based on the provided fine annotated training data. Let's denote the student network and teacher network as $S$ and $T$, respectively. In order to transfer enough informative knowledge from teacher network for learning a robust student network, we formulate the objective function for our student-teacher learning as

\begin{align}
L\;=\;L_S+ r(S,T)  \label{Equ:loss}
\end{align}
where $L_S$ indicates the traditional segmentation (cross entropy) loss for the employed student network. $r(S,T)$ is a function indicating the knowledge bias between the learned student network and teacher network. It serves as a regularization term for regularizing our student learning. In this term, the student and teacher networks are connected together and the knowledge can be distilled from teacher network $T$ to student network $S$ by minimizing $L$. Here, we define $r(S,T)$ as
\begin{align}
r(S,T)=\alpha L_p(S,T)+ \beta L_c(S,T) \label{Equ:lossSub}
\end{align}
%we not only add l2-norm approximate probability distribution loss and l2-norm approximate consistency loss to the loss function, we call them l2-probability $L_P(W_T,W_S)$ and l2-consistency $L_{C}(W_T,W_S)$ respectively that will be explained in the follow section, but also use data distillation to predict the unlabel image with strong teacher model to increase the training samples.The loss function is shown as follow:
%Loss function
%$$Loss\;=\;L_{cross\_entropy}+L_P(W_T,W_S)+L_{C}(W_T,W_S)$where $L_p$ and $L_c$ are two loss functions measuring the information gap between student network and teacher network from two different views. We use the parameter $\gamma$ to control the contribution of the knowledge gained from the teacher network and the student network. Similar to the work of\cite{zagoruyko2016paying} \cite{huang2017like}, we keep the teacher network fixed and optimize over the student network for minimizing the objective function (\ref{Equ:loss}). In the following, we discuss each loss one by one.

$L_p$ is the probability loss defined as $L_p(S,T)=\sum_{i=1,2,...,I} \sum_{\mathbf{x} \in G} \|  \mathbf{p}_S^i(\mathbf{x})-\mathbf{p}_T^i(\mathbf{x}) \|_{2}^2$, where $I$ is the batch size of model's input and $\mathbf{p}_S^i(\mathbf{x}),\mathbf{p}_T^i(\mathbf{x}) \in R^{n}$ are the probability outputs of the student and teacher network at pixel $\mathbf{x}$ in the image region $G$. It is defined in the way such that the probability output of the student network is similar with that of the teacher network. This function can capture the zero-order knowledge between different segmentation outputs.

In complement to $L_p$, term $L_c$ is used to capture the first-order knowledge between outputs of student and teacher network. We formulate it as $L_c(S,T) = \sum_{i=1,2,...,I} \sum_{\mathbf{x} \in G} \|  \mathbf{c}_S^i(\mathbf{x}) -\mathbf{c}_T^i(\mathbf{x})  \|_{2}^2$, \bing{where $I$ is the batch size of model's input and the consistence map $\mathbf{c}$ is defined as $\mathbf{c}_\cdot^i(\mathbf{x})=\sum_{\mathbf{y} \in B (\mathbf{x} )}\|l( \mathbf{y})-l{(\mathbf{x})}\|_2^2$. Here, $B (\mathbf{x} )$ indicates the 8-neighborhood of pixel $\mathbf{x}$ and $l$ is the logits output of the corresponding network}. This term is employed to ensure that the segmented boundary information obtained by student and teacher networks can be closed with each other. By this way, the teacher network provides some useful fist-order knowledge for regularizing our student network learning.

Overall, the above two loss terms (i.e., $L_p$ and $L_c$) constrain the student network learning from different perspectives. They complement well with each other for improving the learning of shallowed student network. Our scheme has the following characteristics for segmentation. \bing{First, it can improve the student segmentation network without incurring extra computational overhead. Second, both the zero-order and first-order knowledge are transferred from teacher to guide our student learning.}

\subsection{Teacher-student learning with unlabeled data}

In addition to the fine annotated images in the training set, we can also obtain a large number of unlabeled images from the Internet for the network training. However, it is unrealistic to annotate all the available images as manually annotating image for segmentation in a pixel-level is quite time consuming. Here, we illustrate that our teacher-student learning framework can be easily extended to make use of these unlabeled images for further improving the learning of student network. In the framework, we treat the segmentation results obtained by the teacher network as the ground truth segmentations for the unlabeled images and then conduct our teacher-student learning on the unlabeled data. Therefore, here we have a total of two teacher-student learning, one is conducted on the manually labeled training set with fine annotations and the other is conducted on the unlabeled data with noisy annotations generated by the teacher network. Both the teacher-student learning can be learned jointly. Specifically, the objective function for our teacher-student learning with both labeled and unlabeled data can be formulated as
\begin{align}
L=L_{Labeled Data}+ \lambda L_{unlabeled Data} \label{Equ:lossUnlabeled}
\end{align}
where $L_{Labeled Data}$ is the loss for the teacher-student learning on the fine annotated training data. $L_{unlabeled Data}$ indicates the loss for the teacher-student learning on the unlabeled data. Here, we use parameter $\lambda$ to control the balance of teacher-student learning for different data. Finally, our teacher-student learning with unlabeled data is achieved by minimizing the loss $L$ defined in (\ref{Equ:lossUnlabeled}).

\section{Experiments}
%We evaluate our method on three challenging datasets, namely Pascal Context\cite{ref31}, PASCAL VOC 2012\cite{ref29}, and Cityscapes\cite{ref30}, each of which is described below.

%In the following, we evaluate our proposed teacher-student learning framework and report the results on three benchmark segmentation sets: Pascal context, Cityscape, and VOC 2012.

\subsection{Ablation study}

In this section, we perform ablation studies on Pascal Context \cite{mottaghi2014role} to justify the effectiveness of our technical contributions in Section \ref{Section:Approach}. We adopt state-of-the-art segmentation architecture DeepLab-v2\cite{chen2018deeplab} as our teacher and student network in the ablation analysis.
In detail, DeepLab-v2 is a stack of two consecutive functional components: (1), a pre-trained ConvNet for local feature extraction (feature backbone network); and (2), Atrous Spatial Pyramid Pooling (ASPP) network for context aggregation. In general, the model capacity of DeepLab-v2 largely correlates with that of feature backbone network. Thus in our ablation experiments, we instantiate our teach network with a higher capacity feature backbone network ResNet-101 \cite{he2016deep}, and employ a recent more computational efficient network MobileNet\cite{howard2017mobilenets} in student network.\footnote{This simply represents a typical teacher-student pair, where teacher is a heavy and accurate segmentation network and student, in the contrary, is an efficient and less-accurate network.}

%In this section, we perform ablation studies \jfh{on Pascal Context \cite{ref31}} to justify the effectiveness of our technical contributions in Section \ref{Section:Approach}. We adopt state-of-the-art \jfh{segmentation architecture} DeepLab-v2\cite{ref2} as our \jfh{teacher and student network} \jfh{in the} ablation analysis. In terms of backbone network in DeepLab-v2, we use ResNet-101\cite{ref39} in teacher network, whereas we use a \bing{recent} \jfh{more computational efficient network MobileNet\cite{ref7}} in student network\footnote{This simply represents a typical teacher-student pair, where teacher is a heavy and accurate segmentation network and student, in the contrary, is an efficient and less-accurate network.}.

\textbf{Dataset:}Pascal Context \cite{mottaghi2014role} has 10103 images, out of which 4998 images are used for training. The images are from Pascal VOC 2010 datasets, and they are annotated as pixelwise segmentation maps which include 540 semantic classes (including the original 20 classes). Each image has approximately $375 \times 500$ pixels. Similar to Mottaghi et al. \cite{mottaghi2014role}, we only consider the most frequent 59 classes in the dataset for evaluation.

\textbf{Implementation Details:} The segmentation networks are trained by batch-based stochastic gradient descent with momentum (0.9). The learning rate is initialized as 0.1, and it is dropped by a factor of 10 after 30, 40 and 50 epoches are reached (60 epoches in total). The images are resized to have maximum length of 512 pixels, and they are zero padded to have square size to allow for batch processing. Besides, general data augmentation methods are used in network training, such as randomly flipping the images, randomly performing scale jitter (scales are between 0.5 to 1.5 ), etc. $\alpha$ and $\beta$ in Equation 2  are empirically set to 4 and 0.4 respectively.we have validate that alpha and beta need to make the probability loss,consistency loss and crose entropy on the same order of magnitude. We randomly take 10k unlabeled images from COCO unlabel 2017 dataset\cite{lin2014microsoft}, and use the teacher segmentation network to generate their pseudo ground truth pixelwise maps. To reduce noises, pixels will not be annotated if their corresponding class likelihood is less confident than 0.7. We implement the proposed network architecture in Tensorflow \cite{abadi2016tensorflow} framework and the algorithms are run on a GPU 1080Ti device.

% we select 10k unlabel data from the COCO dataset which have about 20k unlabel data
%We implement the proposed network architecture on Tensorflow \cite{ref32} framework. During training, each image is randomly mirrorred, scaled (with a scale between 0.5 and 1.5) and random cropping or padding a 512x512 region from the image. The standard stochastic gradient descent with momentum is used for model optimization, where the initial learning rate, momentum and weight decay are set to 0.1, 0.9 and 0.00001, respectively. We initialize the weights of MobileNet-1.0-DeeplabV2 using the MobileNet pre-trained on ILSVRC dataset\cite{ref33}, while the weights for the layers after conv5 are initialized with zero-mean Gaussians. The network converges after about 60K SGD iterations with a mini-batch of 20 samples.

%Learning the knowledge of pixels probability from the teacher network by the probability loss, the student network can

\newcommand{\tabincell}[2]{\begin{tabular}{@{}#1@{}}#2\end{tabular}}
% begin
\begin{table}[t]
\centering
\begin{tabular}{lcc}
\toprule
Model & mIOU($\%$) & speed (FPS) \\
\midrule
 %& \tabincell{c}{Mbd-Lc} & \tabincell{c}{Mbd-Lp} & \tabincell{c}{Mbd-Lc-Lp}& \tabincell{c}{Mbd-Lc-Lp\\-Unlabel} \\ \hline  % \hline
ResNet-101-DeepLab-v2 (teacher) \cite{chen2018deeplab} & 48.5  &16.7 \\ %&38.1 &38.5 &39.2 &40.09
\midrule
MobileNet-1.0-DeepLab-v2    &40.9   &46.5 \\
MobileNet-1.0-DeepLab-v2 (${L}_p$ ) &42.3   &46.5 \\
MobileNet-1.0-DeepLab-v2 (${L}_p$ +${L}_c$ ) & 42.8 &46.5 \\
MobileNet-1.0-DeepLab-v2 (${L}_p$ +${L}_c$+$UnlabeledData$) &43.8 &46.5 \\
\midrule
 FCN-8s \cite{long2015fully} & 37.8 & N/A\\
 ParseNet \cite{liu2015semantic} & 40.4 & N/A\\
 UoA-Context + CRF \cite{lin2016efficient} & 43.3 & < 1 \\
 DAG-RNN \cite{shuai2017scene} & 42.6 & 9.8\\
 DAG-RNN + CRF \cite{shuai2017scene} & 43.7 & < 1 \\
\bottomrule
\end{tabular}
\caption{Comparison results on Pascal Context dataset.  'MobileNet-1.0-DeepLab-v2 (${L}_p$)' indicates that probability loss is considered in the loss for knowledge transfer; 'MobileNet-1.0-DeepLab-v2 (${L}_p$ +${L}_c$)' indicates that probability loss and consistency loss are both used for knowledge transfer; 'MobileNet-1.0-DeepLab-v2 (${L}_p$ +${L}_c$+$UnlabeledData$)' represents that the unlabeled images are used in the knowledge transfer.}
\label{tab:pascalContext}
\end{table}

\begin{figure*}[t]
\begin{center}
%\fbox{\rule{0pt}{2in} \rule{.9\linewidth}{0pt}}
\includegraphics[width=0.7\linewidth, height= 0.4\linewidth]{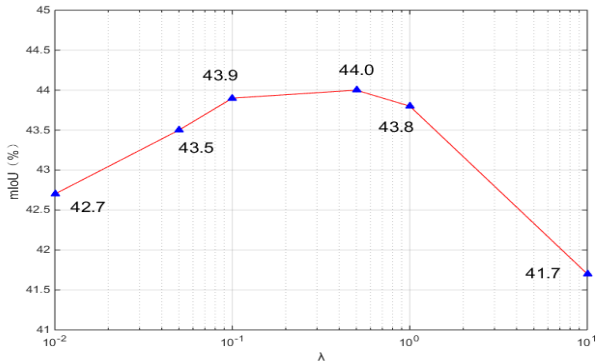}
\end{center}
\vspace{-0.4cm}
   \caption{Evaluation on the influence of parameter $\lambda$.}
\label{fig:Lambda}
\end{figure*}

\textbf{Results:} The results of ablation studies are shown in Table \ref{tab:pascalContext}, where we can observe that the teacher network achieves 48.5$\%$  mIoU \footnote{We note that the reported mIoU (48.5$\%$)  is much higher than that in \cite{chen2018deeplab} (44.7$\%$). This is because that our model has been pre-trained on the MS-COCO dataset. We also find that freezing the batch normalization statistics can benefit our model training significantly, which mainly contributes to the performance superiority.} at 16.7 fps and the student network yields 40.9$\%$ mIoU  at 46.5 fps. Not surprisingly, teacher network significantly outperforms its student counterpart in terms of segmentation accuracy. In contrast, student network can run in real-time, and it has the potential to be applied in real-time application scenario. They are a reasonably appropriate teacher-student setting in our ablation experiments. As expected, the segmentation accuracy of our student network is improved by $1.4$\% to 42.3\% mIOU if we transfer the possibility knowledge from teacher network by only considering ${L}_p$ loss in Equation 2 i.e., the case of $\alpha=4.0, \beta=0$. It demonstrates that the probability distribution output by teacher network indeed carry informative knowledge for improving the learning of our student network.
%(\textcolor{red}{the pixel prediction information in semantic segmentation can be transferred from teacher network to student network}) \textcolor{red}{Bing To Jiafeng and Jianfang: Please explain and elaborate.}
We can observe another promising 0.5\% mIOU gain if the consistency loss ${L}_c$ is further included. This encouraging result indicates that the proposed consistency loss is able to positively guiding the student network learning.
%(\textcolor{red}{the boundary information of the segmentations from teacher network can be transferred to student network and it is complementary to the pixel prediction information})
Finally, when we use the 10K unlabeled images to further facilitate our teacher-student learning, we can observe a significant mIOU improvement (1.0\%). This demonstrates that the knowledge gained by the teacher network can be embedded in unseen data, via which the knowledge is implicitly distilled to the student network.
%(\textcolor{blue}{the knowledge of teacher network can be further transferred to student by using unlabeled images with their pseudo ground truths generated by the teacher network})
Overall, the performance of student has been largely improved after digesting the knowledge from teacher with the proposed teacher-student learning framework. In comparison with state-of-the-arts, our enhanced student networks achieve very competitive results both in terms of inference efficiency as well as segmentation accuracy. We also experimentally find that our system is quite robust to the setting of some parameters like the rate of $L_{unlabeled Data}$ in equation \ref{Equ:lossUnlabeled}. For example, if we set $\lambda$ to 0.1 or 1, the performance will only drop slightly (no more than 0.1\%), which is illustrated in Figure \ref{fig:Lambda}.

\begin{figure*}[t]
\begin{center}
%\fbox{\rule{0pt}{2in} \rule{.9\linewidth}{0pt}}
\includegraphics[width=0.9\linewidth, height=0.6 \linewidth]{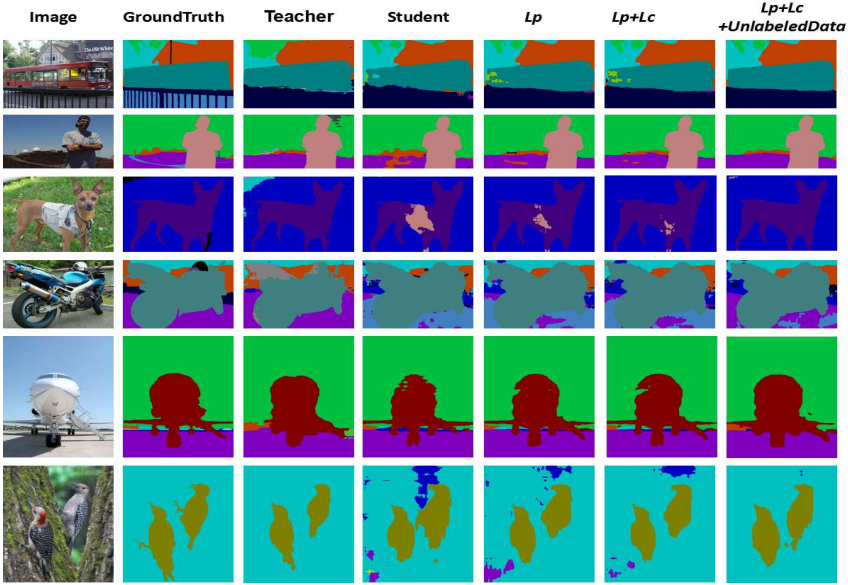}
\end{center}
\vspace{-0.4cm}
   \caption{Some qualitative results on the Pascal Context dataset. The results illustrate that the proposed teacher-student learning framework can efficiently mine informative knowledge from teacher network to guide the learning of student network, and thus improve the performance of student network for segmenting objects. The figure is best viewed in color.}
\label{fig:QuaPascal}
\end{figure*}

In Figure \ref{fig:QuaPascal}, we present several interesting qualitative results. We can easily observe that additionally considering the information bias $r(S,T)$ in student loss function significantly improves the segmentation quality of student network. Specifically, the semantic predictions for "stuff" classes are smoother, and boundaries for "thin" classes (e.g. objects) are slightly shaper. Moreover, those prediction errors can be further decreased when more unlabeled images are incorporated into student network training.
%We can easily observe that \jfh{additionally minimizing the information bias $r(S,T)$} in student loss function significantly improves the quality of the label prediction maps output by student networks. In detail, the semantic predictions for ``stuff" classes are smoother, and boundaries for ``thing" classes (e.g. objects) are slightly shaper. Moreover, those prediction errors can be further decreased when more unlabeled images are incorporated into student network training.}

%end

% \begin{table}[tbp]
% \centering  %
% \setlength{\tabcolsep}{2.0mm}{
% \begin{tabular}{lcccc}
% \hline
% model &DAG-RNN\cite{shuai2016dag} & ResNet-DeepLabV2\cite{chen2018deeplab} & Student &Enhanced Student\\ \hline
% FPS  &9.0  &16.7 &46.5 &46.5\\ \hline
% mIoU &42.6 &48.5 &40.9 &43.8\\ \hline
% \end{tabular}}
% \caption{Comparison with state-of-the-arts on Pascal Context validation set. ALL the inference speeds in this set are evaluated for the segmentation of $500\times N$-sized images, where $N$ ranges from 300 to 500.}
% \label{tab:CompareStatePascalContext}
% \end{table}

% \begin{table}[t]
% \begin{center}
% \begin{tabular}{lcc}
% \toprule
% Model  & mIoU (\%) & FPS \\
% \midrule
% DAG-RNN \cite{shuai2017scene} & 42.6 & 9.8\\
% ResNet-101-DeepLab-v2 \cite{chen2018deeplab} & 48.5 &16.7 \\
% \midrule
% MobileNet-1.0-DeepLab-v2 & 40.9 & 46.5 \\
% MobileNet-1.0-DeepLab-v2  Enhanced& 43.8 & 46.5 \\
% \bottomrule
% \end{tabular}
% \end{center}
% \caption{Comparison with state-of-the-arts on Pascal Context validation set. ALL the inference speeds in this set are evaluated for the segmentation of $500\times N$-sized images, where $N$ ranges from 300 to 500.}
% \label{tab:CompareStatePascalContext}
% \end{table}

\subsection{More evaluations on the student network.}
Given a pair of teacher and student networks, our teacher-student learning framework aims at mining informative knowledges from teacher network to improve the student learning. Here, we provide some experimental clues on how the students affects our teacher-student learning.  Thus, we fix the teacher network, and instantiate three student networks whose performances (segmentation accuracy and speed) are significantly different. Specifically, three different DeepLab-v2 instances with MobileNet-1.0, MobileNet-0.75, and MobileNet-0.5 as the backbone are employed to form the student networks.

%\bing{I'm running out of time, please add something here!!!}

% \begin{table}[t]
% \centering  %
% \begin{tabular}{lccccc|}  % {lccc}left-l,right-r,center-c
% \hline
% Backbone & \tabincell{c}{Teacher} & \tabincell{c}{ Student } & \tabincell{c}{Enhanced Student} \\ \hline  % \hline
% \tabincell{c}{MobileNet-0.5} &48.5 &36.7 &41.1\\ \hline
% \tabincell{c}{MobileNet-0.75} &48.5 &39.2 &43.0\\ \hline
% \tabincell{c}{MobileNet-1.0} &48.5 &40.9 &43.8\\ \hline
% \end{tabular}
% \caption{Comparison results of using different student networks on Pascal Context validation dataset. We use the same teacher network and three different performance student network.}
% \label{tab:EvaStudent}
% \end{table}

\begin{table}[t]
\begin{center}
\begin{tabular}{lccc}
\toprule
Backbone (Student)  & Base(mIOU($\%$)) & Enhanced(mIOU($\%$)) &speed (FPS)\\
\midrule
MobileNet-0.5 & 36.7 & 41.1 &77.7\\
MobileNet-0.75 & 39.2 & 43.0 &61.7\\
MobileNet-1.0 & 40.9 & 43.8 &46.5\\
\bottomrule
\end{tabular}
\end{center}
\caption{Results of using different student networks on Pascal Context validation dataset. The same teacher network (ResNet-101-DeepLab-v2) is used to all three student networks.}
\label{tab:EvaStudent}
\end{table}

%\noindent \textbf{Results:}
The detailed evaluation results are presented in Table \ref{tab:EvaStudent}, where the evaluated three teacher-student network settings have different performance gaps (7.6$\%$, 9.3$\%$ and 11.8$\%$ mIoU). As expected, transferring knowledge from teacher network always improve the performances of students. Specifically, the segmentation accuracies of these three students are non-trivially improved by $2.9$\%, $3.8$\% and $4.4$\% mIOU, respectively. We can observe that the larger performance gaps between the teacher and student is, the more knowledge is gained, and thus the higher performance improvement is observed for the student network.  This observation suggests that the performance of a student network (e.g. MobileNet-1.0-DeepLab-v2) can be further improved by using a stronger teacher model.

\subsection{Comparison with state-of-the-arts}
To further demonstrate the effectiveness of the proposed teacher-student learning approach, we test our methods on Cityscape dataset and Pascal VOC2012 dataset. In the following experiments, we use MobileNet-1.0-DeeplabV2 and ResNet-101-Deeplab-v2 as our student and teacher, respectively. \bing{It's important to note that our learning framework is versatile to different teacher and student models, so the reported results can be simply improved by using either a stronger teacher or a better student model.}

\subsubsection{Experiments on Cityscapes dataset}
%In this part we want to confirm the effect of our proposed knowledge distillation methods for semantic segmentation on Cityscape dataset. We use Resnet-DeeplabV2 as the teacher network and MobileNet-1.0-DeeplabV2 as the student network to compare with the state-of-the-art model on Cityscape dataset.

The Cityscapes\cite{cordts2016cityscapes} is a large-scale dataset for semantic segmentation. This set is captured from the urban streets distributed in 50 cities for the purpose of understanding urban streets. The captured images have a large resolution of $2048 \times 1024$. For evaluation, a total of 5000 images are selected to be annotated in a fine scale and 20000 images are selected to be annotated coarsely. We follow the evaluation protocol in \cite{cordts2016cityscapes}, where 2975, 500, and 1525 of the fine annotated images are selected to train, validate and test the model, respectively.

The detailed comparison results are presented in Table \ref{tab:Cityscapes}. As expected, by employing the proposed teacher-student learning framework, the performance of student network is improved to 71.9\% mIoU, which is about 4.6\% mIoU higher than the original student network. We can also note that the student network enhanced by our teacher-learning algorithm can even perform better than the employed teacher network (71.9 vs. 70.9), which demonstrates the effectiveness of our proposed teacher-student learning framework for mining informative knowledge to guide the learning of student network. We also observe that the enhanced student network can segment images at a fast speed with a good accuracy and it outperforms the SegNet \cite{badrinarayanan2017segnet} in terms of segmentation accuracy and speed.

% \begin{table}[tbp]
% \centering  %
% \setlength{\tabcolsep}{2.0mm}{
% \begin{tabular}{|c|c|c|c|c|c|c|}
% \hline
% model &PSPNet\cite{zhao2017pyramid} & ResNet-DeepLabV2\cite{chen2018deeplab}&SegNet\cite{badrinarayanan2017segnet} & Student &Enhanced Student\\ \hline
% FPS   &6.6 &7.4 &19.6 &20.6 &20.6\\ \hline
% mIoU &80.1&70.9 &56.3 &67.3&71.9\\   \hline
% \end{tabular}}
% \caption{Comparison with state-of-the-arts on Cityscapes validation set. ALL the inference speeds on this set are evaluated for the segmentation of $1024\times512$-sized images.}
% \label{tab:Cityscapes}
% \end{table}

\begin{table}
\begin{center}
\begin{tabular}{lcc}
Model & mIoU (\%) & speed (FPS) \\
\toprule
SegNet \cite{badrinarayanan2017segnet} & 56.3 & 19.6 \\
ResNet-DeepLab-v2 \cite{chen2018deeplab} & 70.9 & 7.4\\
PSPNet \cite{zhao2017pyramid} & \textbf{80.1} & 6.6\\
\midrule
MobileNet-1.0-DeepLab-v2 &  67.3 & 20.6 \\
MobileNet-1.0-DeepLab-v2 (Enhanced) &  71.9 & \textbf{20.6} \\
\bottomrule
\end{tabular}
\caption{Comparison with state-of-the-arts on Cityscapes validation set. ALL the inference speeds on this set are evaluated for the segmentation of $1024\times512$-sized images.}
\label{tab:Cityscapes}
\end{center}
\end{table}

% \begin{table}[tbp]
% \centering  %
% \setlength{\tabcolsep}{2.0mm}{
% \begin{tabular}{|c|c|c|c|c|c|}
% \hline
% model &ICNet\cite{zhao2017icnet} &PSPNet\cite{zhao2017pyramid} & ResNet-DeepLabV2\cite{chen2018deeplab} & Student &Enhanced Student\\ \hline
% FPS  &30.3 &0.35 &0.15 &4.0 &4.0\\ \hline
% mIoU &67.7 &80.1&70.9 &67.25&71.85\\   \hline
% \end{tabular}}
% \caption{Comparison with state-of-the-arts on Cityscapes validation set.}
% \label{tab:Cityscapes}
% \end{table}

\subsubsection{Experiments on Pascal VOC2012}

The Pascal VOC2012 dataset \cite{everingham2012pascal} consists 4369 images of 21 objects classes and a background class. For evaluation, the whole dataset is divided into training, validation, and test sets, each of which has 1446, 1449, and 1456 images, respectively. Following the experiment setup in SDS \cite{hariharan2014simultaneous}, the training set are extended to a set with 10,582.

The detailed comparison results are presented in Table \ref{tab:CompareStateVOC2012}. As can be seen, the student model enhanced by the proposed teacher-student learning framework can obtain a mIoU of 69.6\% on this set, which outperforms the original student network by a margin of 2.3\%. As compared with other state-of-the-arts \cite{zheng2015conditional,yu2015multi}, our enhanced model has large advantage in terms of speed.

\begin{table}[t]
\centering  %
\begin{tabular}{lcc}  % {lccc} left-l,right-r,center-c
\toprule
Model &\tabincell{c}{mIoU($\%$)}  &\tabincell{c}{speed (FPS)}\\
\toprule
%Large Kernel\cite{peng2017large} & 81.0\\ \hline
\tabincell{c}{CRF-RNN\cite{zheng2015conditional}}  & 72.9 &7.6\\
\tabincell{c}{Multi-scale\cite{yu2015multi}}  &73.9 &16.7\\

\tabincell{c}{ResNet-101-DeepLab-v2\cite{chen2018deeplab}}  & \textbf{75.2}  &16.7\\
\midrule
\tabincell{c}{MobileNet-1.0-DeepLab-v2 } &67.3  &46.5\\
\tabincell{c}{MobileNet-1.0-DeepLab-v2 (Enhanced)}  & 69.6 &\textbf{46.5}\\
\bottomrule
\end{tabular}
\caption{Comparison with state-of-the-arts on VOC 2012 validation set.}
\label{tab:CompareStateVOC2012}
\end{table}
% end

\section{Conclusion}
\jfh{In this paper, we have proposed a teacher-student learning framework for improving the performance of existing fast segmentation models. In the framework, both the zero-order and first-order knowledges gained by a teacher network is distilled to regularize our student learning through both fine annotated and unlabeled data. Our experiments show that the proposed learning framework can largely improve the accuracy of student segmentation network without incurring extra computational overhead. The proposed framework mainly mine knowledge from one single teacher network for the student learning. In the future, we would explore multiple teachers based teacher-student learning framework.}

%\vspace{-0.3cm}
\section*{Acknowledgment}
%\vspace{-0.2cm}
This work was supported partially by the NSFC (No. 61702567, 61522115, 61661130157). The corresponding author for this paper is Jian-Fang Hu.

%------------------------------------------------------------------------
%\subsection{Color}
%Color is valuable, and will be visible to readers of the electronic copy. However ensure that, when printed on a monochrome printer, no important information is lost by the conversion to grayscale.
\bibliography{egbib}

\begin{thebibliography}{40}
\providecommand{\natexlab}[1]{#1}
\providecommand{\url}[1]{\texttt{#1}}
\expandafter\ifx\csname urlstyle\endcsname\relax
  \providecommand{\doi}[1]{doi: #1}\else
  \providecommand{\doi}{doi: \begingroup \urlstyle{rm}\Url}\fi

\bibitem[Abadi et~al.(2016)Abadi, Agarwal, Barham, Brevdo, Chen, Citro,
  Corrado, Davis, Dean, Devin, et~al.]{abadi2016tensorflow}
Mart{\'\i}n Abadi, Ashish Agarwal, Paul Barham, Eugene Brevdo, Zhifeng Chen,
  Craig Citro, Greg~S Corrado, Andy Davis, Jeffrey Dean, Matthieu Devin, et~al.
\newblock Tensorflow: Large-scale machine learning on heterogeneous distributed
  systems.
\newblock \emph{arXiv preprint arXiv:1603.04467}, 2016.

\bibitem[Badrinarayanan et~al.(2017)Badrinarayanan, Kendall, and
  Cipolla]{badrinarayanan2017segnet}
Vijay Badrinarayanan, Alex Kendall, and Roberto Cipolla.
\newblock Segnet: A deep convolutional encoder-decoder architecture for image
  segmentation.
\newblock \emph{IEEE transactions on pattern analysis and machine
  intelligence}, 39\penalty0 (12):\penalty0 2481--2495, 2017.

\bibitem[Chen et~al.(2018)Chen, Papandreou, Kokkinos, Murphy, and
  Yuille]{chen2018deeplab}
Liang-Chieh Chen, George Papandreou, Iasonas Kokkinos, Kevin Murphy, and Alan~L
  Yuille.
\newblock Deeplab: Semantic image segmentation with deep convolutional nets,
  atrous convolution, and fully connected crfs.
\newblock \emph{IEEE transactions on pattern analysis and machine
  intelligence}, 40\penalty0 (4):\penalty0 834--848, 2018.

\bibitem[Cordts et~al.(2016)Cordts, Omran, Ramos, Rehfeld, Enzweiler, Benenson,
  Franke, Roth, and Schiele]{cordts2016cityscapes}
Marius Cordts, Mohamed Omran, Sebastian Ramos, Timo Rehfeld, Markus Enzweiler,
  Rodrigo Benenson, Uwe Franke, Stefan Roth, and Bernt Schiele.
\newblock The cityscapes dataset for semantic urban scene understanding.
\newblock In \emph{Proceedings of the IEEE conference on computer vision and
  pattern recognition}, pages 3213--3223, 2016.

\bibitem[Everingham et~al.(2012)Everingham, Van~Gool, Williams, Winn, and
  Zisserman]{everingham2012pascal}
M~Everingham, L~Van~Gool, CKI Williams, J~Winn, and A~Zisserman.
\newblock The pascal visual object classes challenge 2012 (voc2012).
\newblock \emph{Results}, 2012.

\bibitem[G{\"u}l{\c{c}}ehre and Bengio(2016)]{gulccehre2016knowledge}
{\c{C}}aǧlar G{\"u}l{\c{c}}ehre and Yoshua Bengio.
\newblock Knowledge matters: Importance of prior information for optimization.
\newblock \emph{The Journal of Machine Learning Research}, 17\penalty0
  (1):\penalty0 226--257, 2016.

\bibitem[Hariharan et~al.(2014)Hariharan, Arbel{\'a}ez, Girshick, and
  Malik]{hariharan2014simultaneous}
Bharath Hariharan, Pablo Arbel{\'a}ez, Ross Girshick, and Jitendra Malik.
\newblock Simultaneous detection and segmentation.
\newblock In \emph{European Conference on Computer Vision}, pages 297--312.
  Springer, 2014.

\bibitem[He et~al.(2016)He, Zhang, Ren, and Sun]{he2016deep}
Kaiming He, Xiangyu Zhang, Shaoqing Ren, and Jian Sun.
\newblock Deep residual learning for image recognition.
\newblock In \emph{Proceedings of the IEEE conference on computer vision and
  pattern recognition}, pages 770--778, 2016.

\bibitem[Hinton et~al.(2015)Hinton, Vinyals, and Dean]{hinton2015distilling}
Geoffrey Hinton, Oriol Vinyals, and Jeff Dean.
\newblock Distilling the knowledge in a neural network.
\newblock \emph{arXiv preprint arXiv:1503.02531}, 2015.

\bibitem[Howard et~al.(2017)Howard, Zhu, Chen, Kalenichenko, Wang, Weyand,
  Andreetto, and Adam]{howard2017mobilenets}
Andrew~G Howard, Menglong Zhu, Bo~Chen, Dmitry Kalenichenko, Weijun Wang,
  Tobias Weyand, Marco Andreetto, and Hartwig Adam.
\newblock Mobilenets: Efficient convolutional neural networks for mobile vision
  applications.
\newblock \emph{arXiv preprint arXiv:1704.04861}, 2017.

\bibitem[Hu et~al.(2017)Hu, Wang, Shen, van~den Hengel, and
  Porikli]{hu2017pushing}
Qichang Hu, Peng Wang, Chunhua Shen, Anton van~den Hengel, and Fatih Porikli.
\newblock Pushing the limits of deep cnns for pedestrian detection.
\newblock \emph{IEEE Transactions on Circuits and Systems for Video
  Technology}, 2017.

\bibitem[Huang et~al.(2017)Huang, Liu, Weinberger, and van~der
  Maaten]{huang2017densely}
Gao Huang, Zhuang Liu, Kilian~Q Weinberger, and Laurens van~der Maaten.
\newblock Densely connected convolutional networks.
\newblock In \emph{Proceedings of the IEEE conference on computer vision and
  pattern recognition}, volume~1, page~3, 2017.

\bibitem[Huang and Wang(2017)]{huang2017like}
Zehao Huang and Naiyan Wang.
\newblock Like what you like: Knowledge distill via neuron selectivity
  transfer.
\newblock \emph{arXiv preprint arXiv:1707.01219}, 2017.

\bibitem[Kr{\"a}henb{\"u}hl and Koltun(2011)]{krahenbuhl2011efficient}
Philipp Kr{\"a}henb{\"u}hl and Vladlen Koltun.
\newblock Efficient inference in fully connected crfs with gaussian edge
  potentials.
\newblock In \emph{Advances in neural information processing systems}, pages
  109--117, 2011.

\bibitem[Krizhevsky et~al.(2012)Krizhevsky, Sutskever, and
  Hinton]{krizhevsky2012imagenet}
Alex Krizhevsky, Ilya Sutskever, and Geoffrey~E Hinton.
\newblock Imagenet classification with deep convolutional neural networks.
\newblock In \emph{Advances in neural information processing systems}, pages
  1097--1105, 2012.

\bibitem[Lin et~al.(2016)Lin, Shen, Van Den~Hengel, and Reid]{lin2016efficient}
Guosheng Lin, Chunhua Shen, Anton Van Den~Hengel, and Ian Reid.
\newblock Efficient piecewise training of deep structured models for semantic
  segmentation.
\newblock In \emph{Proceedings of the IEEE Conference on Computer Vision and
  Pattern Recognition}, pages 3194--3203, 2016.

\bibitem[Lin et~al.(2017)Lin, Milan, Shen, and Reid]{lin2017refinenet}
Guosheng Lin, Anton Milan, Chunhua Shen, and Ian Reid.
\newblock Refinenet: Multi-path refinement networks for high-resolution
  semantic segmentation.
\newblock In \emph{IEEE Conference on Computer Vision and Pattern Recognition
  (CVPR)}, 2017.

\bibitem[Lin et~al.(2014)Lin, Maire, Belongie, Hays, Perona, Ramanan,
  Doll{\'a}r, and Zitnick]{lin2014microsoft}
Tsung-Yi Lin, Michael Maire, Serge Belongie, James Hays, Pietro Perona, Deva
  Ramanan, Piotr Doll{\'a}r, and C~Lawrence Zitnick.
\newblock Microsoft coco: Common objects in context.
\newblock In \emph{European conference on computer vision}, pages 740--755.
  Springer, 2014.

\bibitem[Liu et~al.(2015{\natexlab{a}})Liu, Rabinovich, and
  Berg]{liu2015parsenet}
Wei Liu, Andrew Rabinovich, and Alexander~C Berg.
\newblock Parsenet: Looking wider to see better.
\newblock \emph{arXiv preprint arXiv:1506.04579}, 2015{\natexlab{a}}.

\bibitem[Liu et~al.(2015{\natexlab{b}})Liu, Li, Luo, Loy, and
  Tang]{liu2015semantic}
Ziwei Liu, Xiaoxiao Li, Ping Luo, Chen-Change Loy, and Xiaoou Tang.
\newblock Semantic image segmentation via deep parsing network.
\newblock In \emph{Computer Vision (ICCV), 2015 IEEE International Conference
  on}, pages 1377--1385. IEEE, 2015{\natexlab{b}}.

\bibitem[Long et~al.(2015)Long, Shelhamer, and Darrell]{long2015fully}
Jonathan Long, Evan Shelhamer, and Trevor Darrell.
\newblock Fully convolutional networks for semantic segmentation.
\newblock In \emph{Proceedings of the IEEE conference on computer vision and
  pattern recognition}, pages 3431--3440, 2015.

\bibitem[Mottaghi et~al.(2014)Mottaghi, Chen, Liu, Cho, Lee, Fidler, Urtasun,
  and Yuille]{mottaghi2014role}
Roozbeh Mottaghi, Xianjie Chen, Xiaobai Liu, Nam~Gyu Cho, Seong~Whan Lee, Sanja
  Fidler, Raquel Urtasun, and Alan Yuille.
\newblock The role of context for object detection and semantic segmentation in
  the wild.
\newblock In \emph{Proceedings of the IEEE Conference on Computer Vision and
  Pattern Recognition}, pages 891--898, 2014.

\bibitem[Noh et~al.(2015)Noh, Hong, and Han]{noh2015learning}
Hyeonwoo Noh, Seunghoon Hong, and Bohyung Han.
\newblock Learning deconvolution network for semantic segmentation.
\newblock In \emph{Proceedings of the IEEE International Conference on Computer
  Vision}, pages 1520--1528, 2015.

\bibitem[Paszke et~al.(2016)Paszke, Chaurasia, Kim, and
  Culurciello]{paszke2016enet}
Adam Paszke, Abhishek Chaurasia, Sangpil Kim, and Eugenio Culurciello.
\newblock Enet: A deep neural network architecture for real-time semantic
  segmentation.
\newblock \emph{arXiv preprint arXiv:1606.02147}, 2016.

\bibitem[Romero et~al.(2014)Romero, Ballas, Kahou, Chassang, Gatta, and
  Bengio]{romero2014fitnets}
Adriana Romero, Nicolas Ballas, Samira~Ebrahimi Kahou, Antoine Chassang, Carlo
  Gatta, and Yoshua Bengio.
\newblock Fitnets: Hints for thin deep nets.
\newblock \emph{arXiv preprint arXiv:1412.6550}, 2014.

\bibitem[Ros et~al.(2016)Ros, Stent, Alcantarilla, and
  Watanabe]{ros2016training}
German Ros, Simon Stent, Pablo~F Alcantarilla, and Tomoki Watanabe.
\newblock Training constrained deconvolutional networks for road scene semantic
  segmentation.
\newblock \emph{arXiv preprint arXiv:1604.01545}, 2016.

\bibitem[Russakovsky et~al.(2015)Russakovsky, Deng, Su, Krause, Satheesh, Ma,
  Huang, Karpathy, Khosla, Bernstein, et~al.]{russakovsky2015imagenet}
Olga Russakovsky, Jia Deng, Hao Su, Jonathan Krause, Sanjeev Satheesh, Sean Ma,
  Zhiheng Huang, Andrej Karpathy, Aditya Khosla, Michael Bernstein, et~al.
\newblock Imagenet large scale visual recognition challenge.
\newblock \emph{International Journal of Computer Vision}, 115\penalty0
  (3):\penalty0 211--252, 2015.

\bibitem[Sharma et~al.(2014)Sharma, Tuzel, and Liu]{sharma2014recursive}
Abhishek Sharma, Oncel Tuzel, and Ming-Yu Liu.
\newblock Recursive context propagation network for semantic scene labeling.
\newblock In \emph{Advances in Neural Information Processing Systems}, pages
  2447--2455, 2014.

\bibitem[Shuai et~al.(2016)Shuai, Zuo, Wang, and Wang]{shuai2016dag}
Bing Shuai, Zhen Zuo, Bing Wang, and Gang Wang.
\newblock Dag-recurrent neural networks for scene labeling.
\newblock In \emph{Proceedings of the IEEE conference on computer vision and
  pattern recognition}, pages 3620--3629, 2016.

\bibitem[Shuai et~al.(2017)Shuai, Zuo, Wang, and Wang]{shuai2017scene}
Bing Shuai, Zhen Zuo, Bing Wang, and Gang Wang.
\newblock Scene segmentation with dag-recurrent neural networks.
\newblock \emph{IEEE transactions on pattern analysis and machine
  intelligence}, 2017.

\bibitem[Simonyan and Zisserman(2014)]{simonyan2014very}
Karen Simonyan and Andrew Zisserman.
\newblock Very deep convolutional networks for large-scale image recognition.
\newblock \emph{arXiv preprint arXiv:1409.1556}, 2014.

\bibitem[Szegedy et~al.(2015)Szegedy, Liu, Jia, Sermanet, Reed, Anguelov,
  Erhan, Vanhoucke, Rabinovich, et~al.]{szegedy2015going}
Christian Szegedy, Wei Liu, Yangqing Jia, Pierre Sermanet, Scott Reed, Dragomir
  Anguelov, Dumitru Erhan, Vincent Vanhoucke, Andrew Rabinovich, et~al.
\newblock Going deeper with convolutions.
\newblock Cvpr, 2015.

\bibitem[Tai et~al.(2016)Tai, Yang, Zhang, Luo, Qian, and Chen]{tai2016face}
Ying Tai, Jian Yang, Yigong Zhang, Lei Luo, Jianjun Qian, and Yu~Chen.
\newblock Face recognition with pose variations and misalignment via orthogonal
  procrustes regression.
\newblock \emph{IEEE Transactions on Image Processing}, 25\penalty0
  (6):\penalty0 2673--2683, 2016.

\bibitem[Wang et~al.(2017)Wang, Chen, Yuan, Liu, Huang, Hou, and
  Cottrell]{wang2017understanding}
Panqu Wang, Pengfei Chen, Ye~Yuan, Ding Liu, Zehua Huang, Xiaodi Hou, and
  Garrison Cottrell.
\newblock Understanding convolution for semantic segmentation.
\newblock \emph{arXiv preprint arXiv:1702.08502}, 2017.

\bibitem[Yim et~al.(2017)Yim, Joo, Bae, and Kim]{yim2017gift}
Junho Yim, Donggyu Joo, Jihoon Bae, and Junmo Kim.
\newblock A gift from knowledge distillation: Fast optimization, network
  minimization and transfer learning.
\newblock In \emph{The IEEE Conference on Computer Vision and Pattern
  Recognition (CVPR)}, 2017.

\bibitem[Yu and Koltun(2015)]{yu2015multi}
Fisher Yu and Vladlen Koltun.
\newblock Multi-scale context aggregation by dilated convolutions.
\newblock \emph{arXiv preprint arXiv:1511.07122}, 2015.

\bibitem[Zagoruyko and Komodakis(2016)]{zagoruyko2016paying}
Sergey Zagoruyko and Nikos Komodakis.
\newblock Paying more attention to attention: Improving the performance of
  convolutional neural networks via attention transfer.
\newblock \emph{arXiv preprint arXiv:1612.03928}, 2016.

\bibitem[Zhao et~al.(2017{\natexlab{a}})Zhao, Qi, Shen, Shi, and
  Jia]{zhao2017icnet}
Hengshuang Zhao, Xiaojuan Qi, Xiaoyong Shen, Jianping Shi, and Jiaya Jia.
\newblock Icnet for real-time semantic segmentation on high-resolution images.
\newblock \emph{arXiv preprint arXiv:1704.08545}, 2017{\natexlab{a}}.

\bibitem[Zhao et~al.(2017{\natexlab{b}})Zhao, Shi, Qi, Wang, and
  Jia]{zhao2017pyramid}
Hengshuang Zhao, Jianping Shi, Xiaojuan Qi, Xiaogang Wang, and Jiaya Jia.
\newblock Pyramid scene parsing network.
\newblock In \emph{IEEE Conf. on Computer Vision and Pattern Recognition
  (CVPR)}, pages 2881--2890, 2017{\natexlab{b}}.

\bibitem[Zheng et~al.(2015)Zheng, Jayasumana, Romera-Paredes, Vineet, Su, Du,
  Huang, and Torr]{zheng2015conditional}
Shuai Zheng, Sadeep Jayasumana, Bernardino Romera-Paredes, Vibhav Vineet,
  Zhizhong Su, Dalong Du, Chang Huang, and Philip~HS Torr.
\newblock Conditional random fields as recurrent neural networks.
\newblock In \emph{Proceedings of the IEEE International Conference on Computer
  Vision}, pages 1529--1537, 2015.

\end{thebibliography}
\end{document}